\documentclass[10pt,twocolumn,letterpaper]{article}

\usepackage[pagenumbers]{cvpr} % To force page numbers, e.g. for an arXiv version

\usepackage{graphicx}
\usepackage{amsmath}
\usepackage{amssymb}
\usepackage{booktabs}

\usepackage[pagebackref,breaklinks,colorlinks]{hyperref}

\usepackage[capitalize]{cleveref}
\crefname{section}{Sec.}{Secs.}
\Crefname{section}{Section}{Sections}
\Crefname{table}{Table}{Tables}
\crefname{table}{Tab.}{Tabs.}

\def\cvprPaperID{*****} % *** Enter the CVPR Paper ID here
\def\confName{CVPR}
\def\confYear{2022}

\begin{document}

%%%%%%%%% TITLE - PLEASE UPDATE
\title{Zero123++: a Single Image to Consistent Multi-view Diffusion Base Model}

\author{Ruoxi Shi$^1$ \quad Hansheng Chen$^2$ \quad Zhuoyang Zhang$^3$ \quad Minghua Liu$^1$ \\
Chao Xu$^4$ \quad Xinyue Wei$^1$ \quad Linghao Chen$^5$ \quad Chong Zeng$^5$ \quad Hao Su$^1$ \\
\\
$^1$UC San Diego \quad $^2$Stanford University \quad $^3$Tsinghua University \quad $^4$UCLA \quad $^5$Zhejiang University
% {\tt\small \{r5shi, \}@ucsd.edu}
% For a paper whose authors are all at the same institution,
% omit the following lines up until the closing ``}''.
% Additional authors and addresses can be added with ``\and'',
% just like the second author.
% To save space, use either the email address or home page, not both
% \and
% Second Author
% Institution2\\
% First line of institution2 address\\
% {\tt\small secondauthor@i2.org}
}

\twocolumn[{
\maketitle
\begin{center}
    \includegraphics[width=0.99\linewidth]{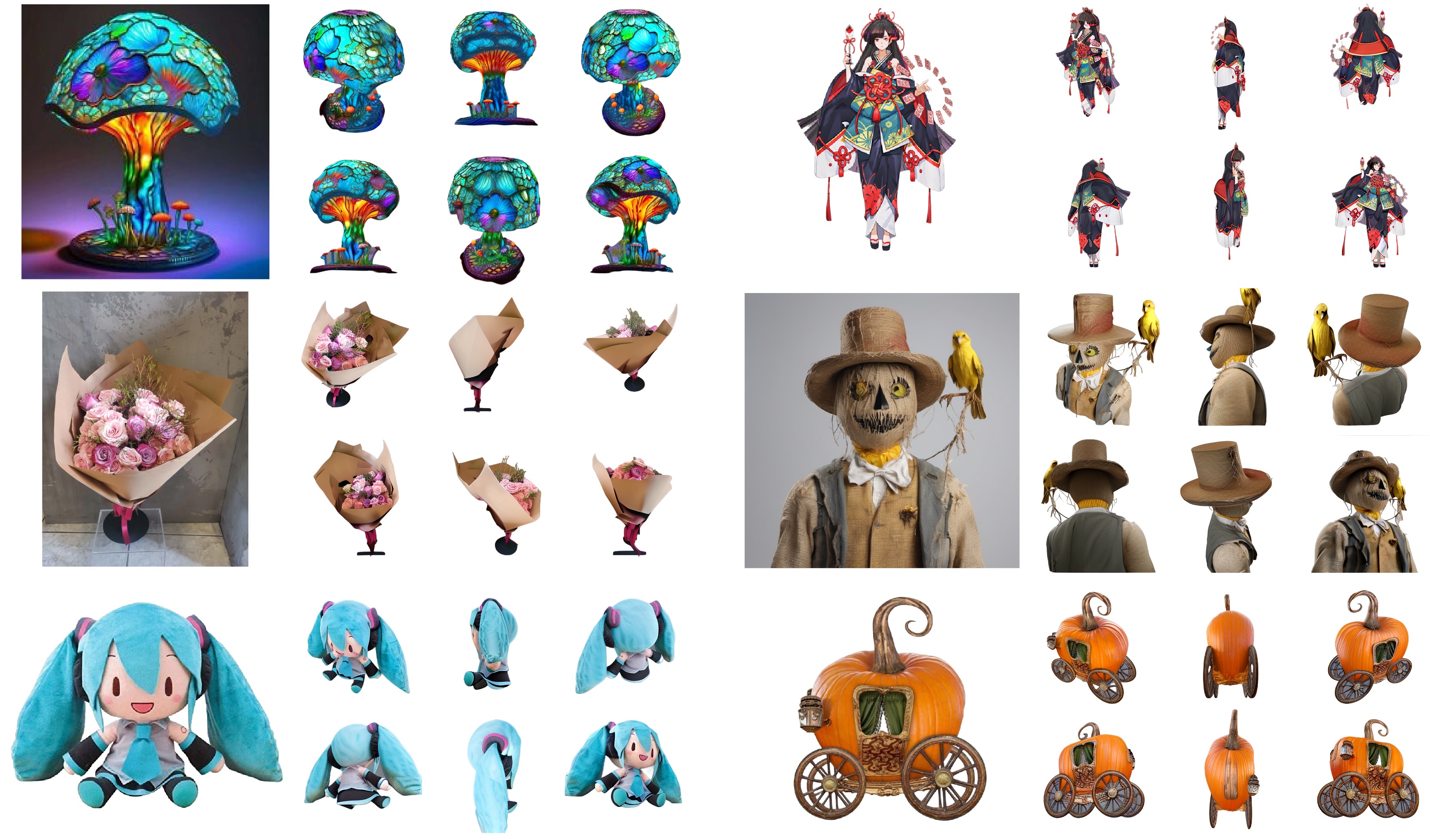}
    \label{fig:teaser}
    \captionof{figure}{Zero123++ excels at generating high-quality, consistent multi-view 3D images, accommodating a wide range of open-world inputs, from real photos to generated images and 2D illustrations.}
    \vspace{0.3cm}
\end{center}
}]

%%%%%%%%% ABSTRACT
\begin{abstract}
   We report Zero123++, 
   % a single image to consistent multi-view diffusion base model.
   an image-conditioned diffusion model for generating 3D-consistent multi-view images from a single input view. 
   To take full advantage of pretrained 2D generative priors, we develop various conditioning and training schemes to minimize the effort of finetuning from off-the-shelf image diffusion models such as StableDiffusion. 
   % enabling making full advantage of powerful 2D pretraining priors.
   % Zero123++ generates consistent, high-quality multi-view images from a single input image,
   % and does not suffer from common problems like degraded texture and geometric misalignment.
   Zero123++ excels in producing high-quality, consistent multi-view images from a single image, overcoming common issues like texture degradation and geometric misalignment. 
   Furthermore, we showcase the feasibility of training a ControlNet on Zero123++ for enhanced control over the generation process.
   The code is available at \url{https://github.com/SUDO-AI-3D/zero123plus}.
\end{abstract}

%%%%%%%%% BODY TEXT
\section{Introduction}
\label{sec:intro}

3D content generation has seen significant progress with the emerging novel view generative models, leveraging the powerful 2D diffusion generative priors learned from extensive datasets sourced from the Internet.
Zero-1-to-3~\cite{liu2023zero} (or Zero123) pioneers open-world single-image-to-3D conversion through zero-shot novel view synthesis. Despite promising performance, the geometric inconsistency in its generated images has yet to bridge the gap between multi-view images and 3D scenes.
%, and shows promising performance.
% However, multi-view images are not 3D yet; consistent ones are.
Recent works like One-2-3-45~\cite{liu2023one}, SyncDreamer~\cite{liu2023syncdreamer} and Consistent123~\cite{lin2023consistent123} build extra layers upon Zero-1-to-3 to obtain more 3D-consistent results. Optimization-based methods like DreamFusion~\cite{poole2022dreamfusion}, ProlificDreamer~\cite{wang2023prolificdreamer} and DreamGaussian~\cite{tang2023dreamgaussian} distill a 3D representation from inconsistent models to obtain 3D results. While these techniques are effective, they could work even better with a base diffusion model that generates consistent multi-view images. In this light, we revisit Zero-1-to-3 and finetune a new multi-view consistent base diffusion model from Stable Diffusion~\cite{Rombach_2022_CVPR}.

% Zero-1-to-3 generates each novel view independently.
% Due to the sampling nature of diffusion models, Zero-1-to-3 usually breaks consistency between generated views.
% To solve this problem, we tile 6 views surrounding the object into one single image as our target for diffusion to correctly model the joint distribution of multi-view images of an object.

Zero-1-to-3 generates each novel view independently. Due to the sampling nature of diffusion models, this approach leads to a breakdown in consistency between the generated views. To address this issue, we adopt a strategy of tiling six views surrounding the object into a single image. This tiling layout enables the correct modeling of the joint distribution of multi-view images of an object.

% Another problem is that Zero-1-to-3 is not making full use of the capabilities of Stable Diffusion.
% We attribute this to two issues in its design: a) when training on image conditions, Zero-1-to-3 did not reuse the existing global or local conditioning mechanism in Stable Diffusion. In Zero123++, we carefully designed various conditioning techniques to maximize reusing Stable Diffusion priors. b) Zero-1-to-3 adopts a reduced resolution for training. It is widely known that the image generation quality of Stable Diffusion models quickly degrades when decreasing output resolution to lower than training resolution. However, the authors of Zero-1-to-3 noticed that training was unstable at the native resolution of 512, and use a low resolution of 256 for training. We analyzed this behavior and proposed a set of strategies to mitigate this issue.

Another issue with Zero-1-to-3 is its underutilization of existing capabilities offered by Stable Diffusion. We attribute this to two design problems:
a) During training with image conditions, Zero-1-to-3 does not effectively incorporate the global or local conditioning mechanisms provided by Stable Diffusion. In Zero123++, we have taken a careful approach, implementing various conditioning techniques to maximize the utilization of Stable Diffusion priors.
b) Zero-1-to-3 uses a reduced resolution for training. It is widely recognized that reducing the output resolution below the training resolution can lead to a decline in image generation quality for Stable Diffusion models. However, the authors of Zero-1-to-3 encountered instability when training at the native resolution of 512 and opted for a lower resolution of 256. We have conducted an in-depth analysis of this behavior and proposed a series of strategies to address this issue.

\section{Improving Consistency and Conditioning}
\label{sec:method}

% In this section, we delve into the techniques used in Zero123++ to enhance multi-view consistency and image conditioning, 
In this section, we explore the techniques employed in Zero123++ to improve multi-view consistency and image conditioning, 
with a primary focus on reusing the priors from pretrained Stable Diffusion model.

\subsection{Multi-view Generation}

\begin{figure}
    \centering
    \includegraphics[width=0.8\linewidth]{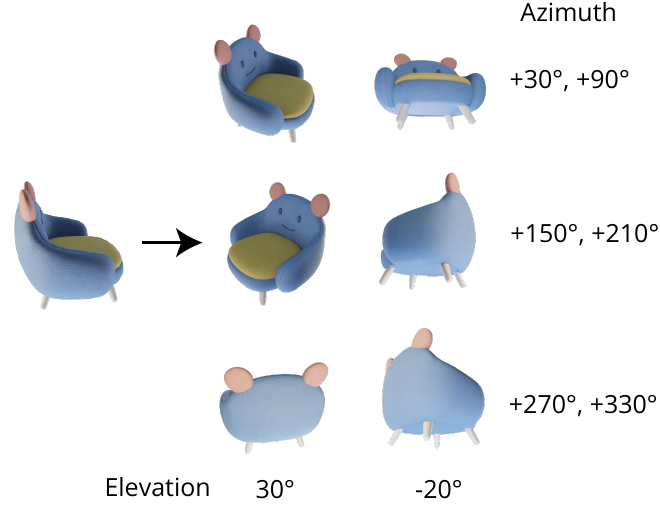}
    \caption{\textbf{Layout of Zero123++ prediction target.} We use a fixed set of relative azimuth and absolute elevation angles.}
    \label{fig:layout}
\end{figure}

The essence of generating consistent multi-view images is the correct modeling of the joint distribution of multiple images. Zero-1-to-3 models the conditional marginal distribution of each image separately and independently, which ignores the correlations between multi-view images.

In Zero123++, we take the simplest form of tiling $6$ images with a $3\times 2$ layout into a single frame for multi-view generation (Fig.~\ref{fig:layout}).

In the context of object and camera poses, it's worth noting that the Objaverse dataset~\cite{deitke2023objaverse} does not consistently align objects in a canonical pose, despite them typically being oriented along the gravity axis. Consequently, there is a wide range of absolute orientations for objects. We have observed that training the model on absolute camera poses can lead to difficulties in disambiguating object orientations.
% We observe that the model suffers from unknown input image azimuth when trained on absolute target poses.
Conversely, Zero-1-to-3 is trained on camera poses with relative azimuth and elevation angles to the input view. This formulation, however, requires knowing the elevation angle of the input view to determine the relative poses between novel views. As a result, various existing pipelines like One-2-3-45~\cite{liu2023one} and DreamGaussian~\cite{tang2023dreamgaussian} have incorporated an additional elevation estimation module, which introduces extra error into the pipeline. 

To address these issues, we use fixed absolute elevation angles and relative azimuth angles as the novel view poses, eliminating the orientation ambiguity without requiring additional elevation estimation. More specifically,  the six poses consist of interleaving elevations of $30^{\circ}$ downward and $20^{\circ}$ upward, combined with azimuths that start at $30^{\circ}$ and increase by $60^{\circ}$ for each pose.

% \subsection{Consistency: the Noise Schedule}
\subsection{Consistency and Stability: Noise Schedule}

\begin{figure}
    \subfloat[]{\includegraphics[width=0.48\linewidth]{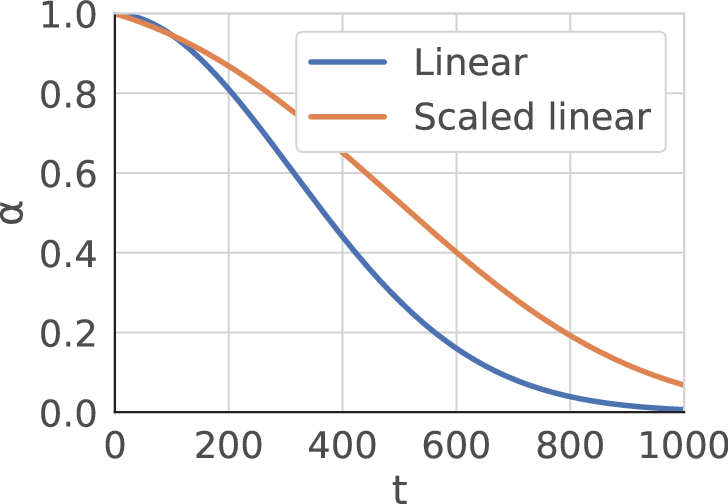}}
    \hfil
    \subfloat[]{\includegraphics[width=0.48\linewidth]{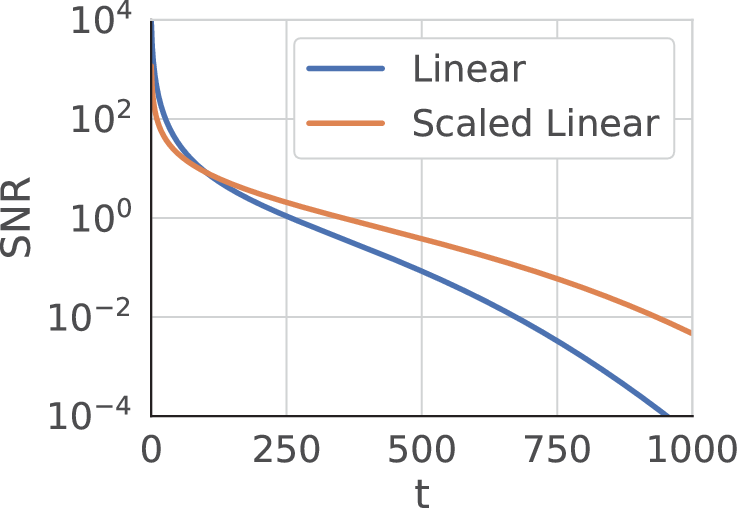}}
    \caption{Comparison between the linear schedule and Stable Diffusion's scaled linear schedule.}
    \label{fig:noisesched}
\end{figure}

The original noise schedule for Stable Diffusion, \ie, the scaled-linear schedule, places emphasis on local details but has very few steps with lower Signal-to-Noise Ratio (SNR), as shown in Fig.~\ref{fig:noisesched}. These low SNR steps occur in the early denoising stage, which is crucial for determining the global low-frequency structure of the content. A reduced number of steps in this stage, either during training or inference, can lead to greater structural variation.
While this setup is suitable for single-image generation, we have observed that it limits the model's ability to ensure the global consistency between multiple views.

\begin{figure}
    \centering
    \includegraphics[width=0.98\linewidth]{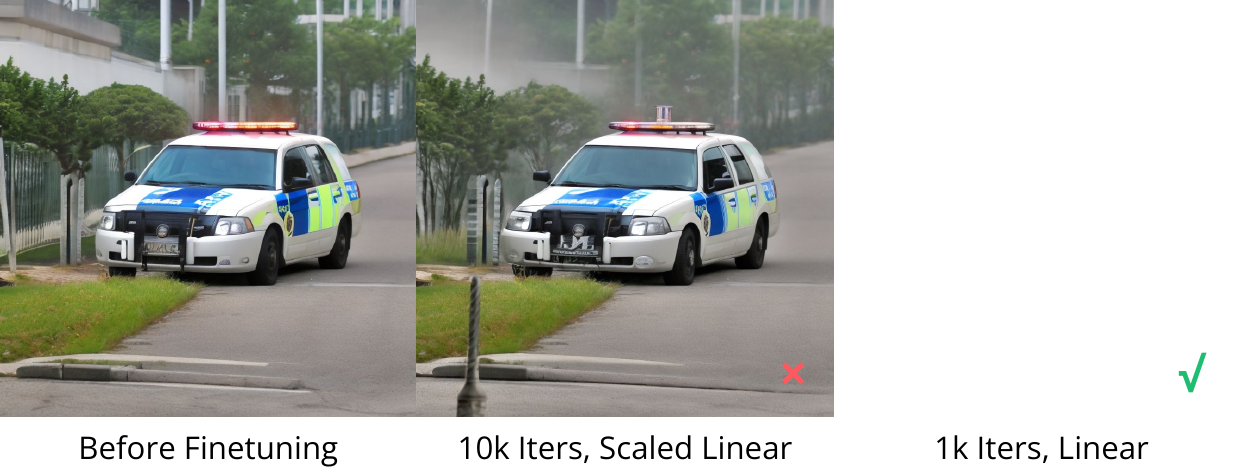}
    \caption{\textbf{Importance of the noise schedule.} Noise schedule strongly affects the model's capability to adapt to new global requirements (generating a pure white image from the prompt \textit{a police car} in this case). Notably, both schedules produce highly similar images before fine-tuning; therefore, we present only the result of the $v$ model with linear schedule before fine-tuning.}
    \label{fig:white}
\end{figure}

To empirically verify this, we perform a toy task by finetuning a LoRA~\cite{hu2021lora} model on the Stable Diffusion 2 v-prediction model to overfit a blank white image given the prompt \textit{a police car}. The results are presented in Fig.~\ref{fig:white}.
Surprisingly, with the scaled-linear noise schedule, the LoRA model cannot overfit on this simple task; it only slightly whitened the image. In contrast, with the linear noise schedule, the LoRA model successfully generates a blank white image regardless of the prompt. While finetuning the full model may still be viable for the scaled-linear schedule, this example highlights the significant impact of the noise schedule on the model's ability to adapt to new global requirements. 

% \begin{table*}
% \centering
% \caption{Table}
% \small
% \begin{tabular}{lcccc} 
% \toprule
% Noise Schedule & Weighting & Parameterization            & LoRA Rank & Iterations to Pure White  \\ 
% \midrule
% Scaled Linear  & SNR             & $\epsilon$ & 4         & Fail ($>$10k)                \\
% Scaled Linear  & 1+SNR           & $v$                       & 4         & Fail ($>$10k)                \\
% Scaled Linear  & Min-SNR~\cite{hang2023efficient}         & $v$                       & 128       & Fail ($>$10k)                \\
% Linear         & 1+SNR           & $v$                       & 4         & 300                       \\
% Linear         & Min-SNR~\cite{hang2023efficient}         & $v$                       & 128       & 200                       \\
% \bottomrule
% \end{tabular}
% \end{table*}

As pointed out by Chen~\cite{chen2023importance}, high-resolution images appear less noisy compared to low-resolution images when subjected to the same absolute level of independent noise (see Fig.~2 in \cite{chen2023importance}).
This phenomenon occurs because ``higher resolution natural images tend to exhibit higher degree of redundancy in (nearby) pixels, therefore less information is destroyed with the same level of independent noise''.
Consequently, we can interpret the use of lower resolution in Zero-1-to-3 training as a modification of the noise schedule, placing greater emphasis on the global requirements of 3D-consistent multi-view generation. 
This also explains the instability issue of training Zero-1-to-3 with higher resolution~\cite{liu2023zero}.

\begin{figure}
    \centering
    \includegraphics[width=0.8\linewidth]{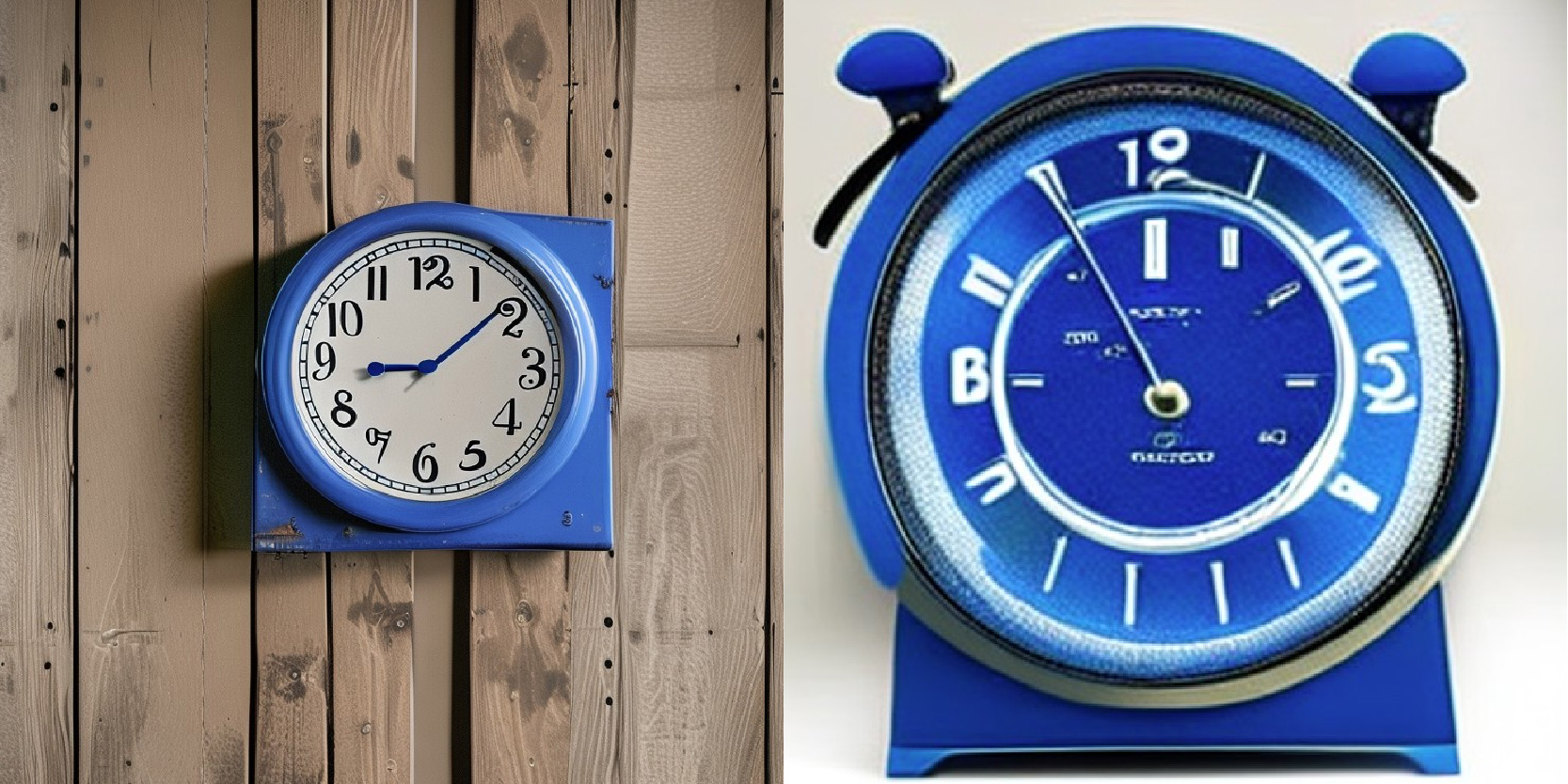}
    \caption{\textbf{Swapping the noise schedule.} We swap the schedule of Stable Diffusion 2 $v$ (Left) and $\epsilon$-parameterized (Right) models from scaled-linear to linear at inference time without any finetuning. Prompt: \textit{a blue clock with black numbers}. The $\epsilon$-parameterized model exhibits a significant decrease in quality, while the $v$ model produces a high-quality image with the linear noise schedule.}
    \label{fig:v-vs-eps}
\end{figure}

In summary, we find it necessary to switch from the scaled-linear schedule to the linear schedule for noise in our model. 
However, this shift introduces another potential challenge: adapting the pretrained model to the new schedule. 
% Fortunately, the $v$-parameterization of diffusion models possesses the unique property of being independent of the noise levels~\cite{salimans2022progressive}, in contrast to the $x_0$- and $\epsilon$-parameterizations.
Fortunately, we have observed that the $v$-prediction model is quite robust when it comes to swapping the schedule, in contrast to the $x_0$- and $\epsilon$-parameterizations, as illustrated in Figure~\ref{fig:v-vs-eps}. It is also theoretically supported that the $v$-prediction is inherently more stable~\cite{salimans2022progressive}. Therefore, we have opted to utilize the Stable Diffusion 2 $v$-prediction model as our base model for fine-tuning.

\subsection{Local Condition: Scaled Reference Attention}

In Zero-1-to-3, the conditioning image (single view input) is concatenated in the feature dimension with the noisy inputs to be denoised for local image conditioning.
This imposes an incorrect pixel-wise spatial correspondence between the input and the target image.

% todo: figure of SNRA pipeline
\begin{figure}
    \centering
    \includegraphics[width=0.82\linewidth]{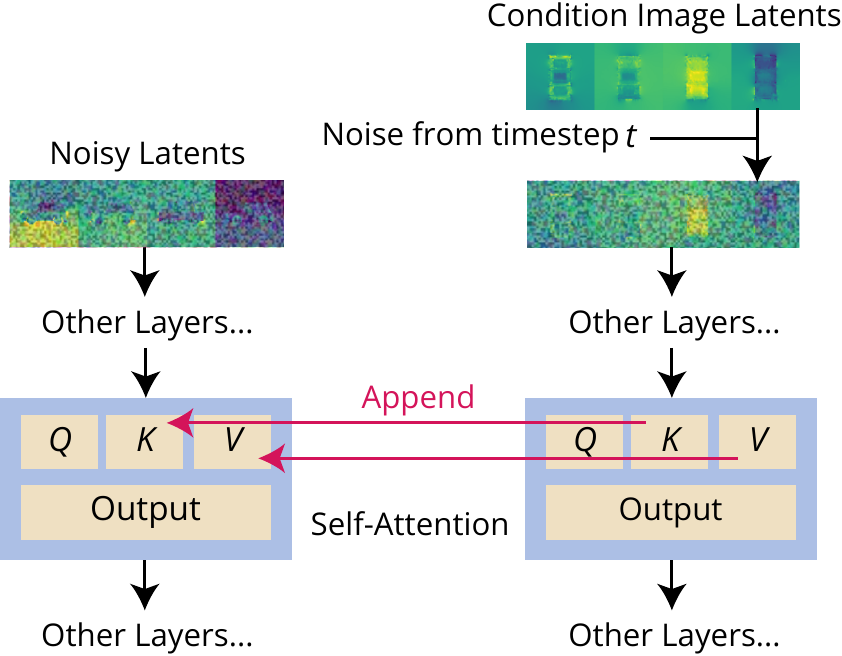}
    \caption{\textbf{Reference Attention.} It adds an additional conditioning branch and modifies key (K) and value (V) matrices of the self-attention layers to accept the extra condition image, which can fully reuse Stable Diffusion priors.}
    \label{fig:ref-attn}
\end{figure}

We propose to use a scaled version of Reference Attention to provide proper local conditioning input.

As shown in Fig.~\ref{fig:ref-attn}, Reference Attention~\cite{refattn} refers to the operation of running the denoising UNet model on an extra reference image and appending the self-attention key and value matrices from the reference image to the corresponding attention layers when denoising the model input. The same level of Gaussian noise as the denoising input is added to the reference image to allow the UNet to attend to relevant features for denoising at the current noise level.

\begin{figure}
    \centering
    \includegraphics[width=\linewidth]{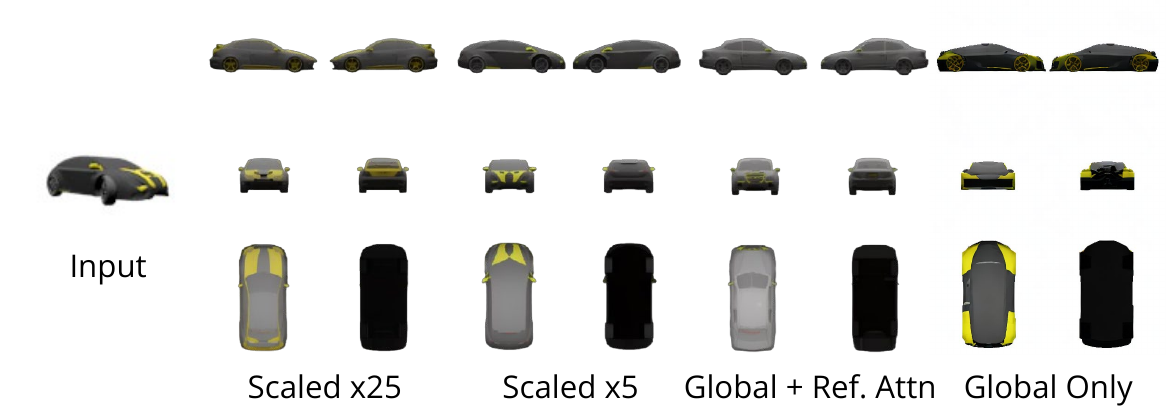}
    \caption{\textbf{Comparison on local conditioning.} We train Zero123++ with different levels of scaled reference attention on the ShapeNet Cars dataset. Output coherence with the input image is best on 5x scaled reference attention.}
    \label{fig:local}
\end{figure}

Without any finetuning, Reference Attention is already capable of guiding the diffusion model to generate images that share similar semantic content and texture with the reference image. When finetuned, we observed that the Reference Attention works better when we scale the latent (before adding noise). In Figure~\ref{fig:local}, we provide a comparison from experiments conducted on ShapeNet Cars~\cite{chang2015shapenet} to demonstrate that the model achieves the highest consistency with the conditioning image when the reference latent is scaled by a factor of 5.

\subsection{Global Condition: FlexDiffuse}

In the original Stable Diffusion, global conditioning comes solely from text embeddings.
Stable Diffusion employs CLIP~\cite{radford2021learning} as the text encoder, and performs cross-attention between model latents and per-token CLIP text embeddings.
As a result, we can make use of the alignment between CLIP image and text spaces to reuse the prior for global image conditioning.

We propose a trainable variant of the linear guidance mechanism introduced in FlexDiffuse~\cite{flexdiffuse} to incorporate global image conditioning into the model while minimizing the extent of fine-tuning. We start from the original prompt embeddings $T$ of shape $L \times D$ where $L$ is length of tokens and $D$ is the dimension of token embeddings, and add the CLIP global image embedding $I$ of shape $D$ multiplied by a trainable set of global weights $\{w_i\}_{i = 1, \dots, L}$ (a shared set of weights for all tokens) to the original prompt embeddings, or formally,

\begin{equation}
    T_i' = T_i + w_i \cdot I, i = 1, 2, \dots, L.
\end{equation}

We initialize the weights with FlexDiffuse's linear guidance:

\begin{equation}
    w_i = \frac{i}{L}.
\end{equation}

In the released Zero123++ models, we do not impose any text conditions, so $T$ is obtained by encoding an empty prompt.

\begin{figure}
    \centering
    \includegraphics[width=0.95\linewidth]{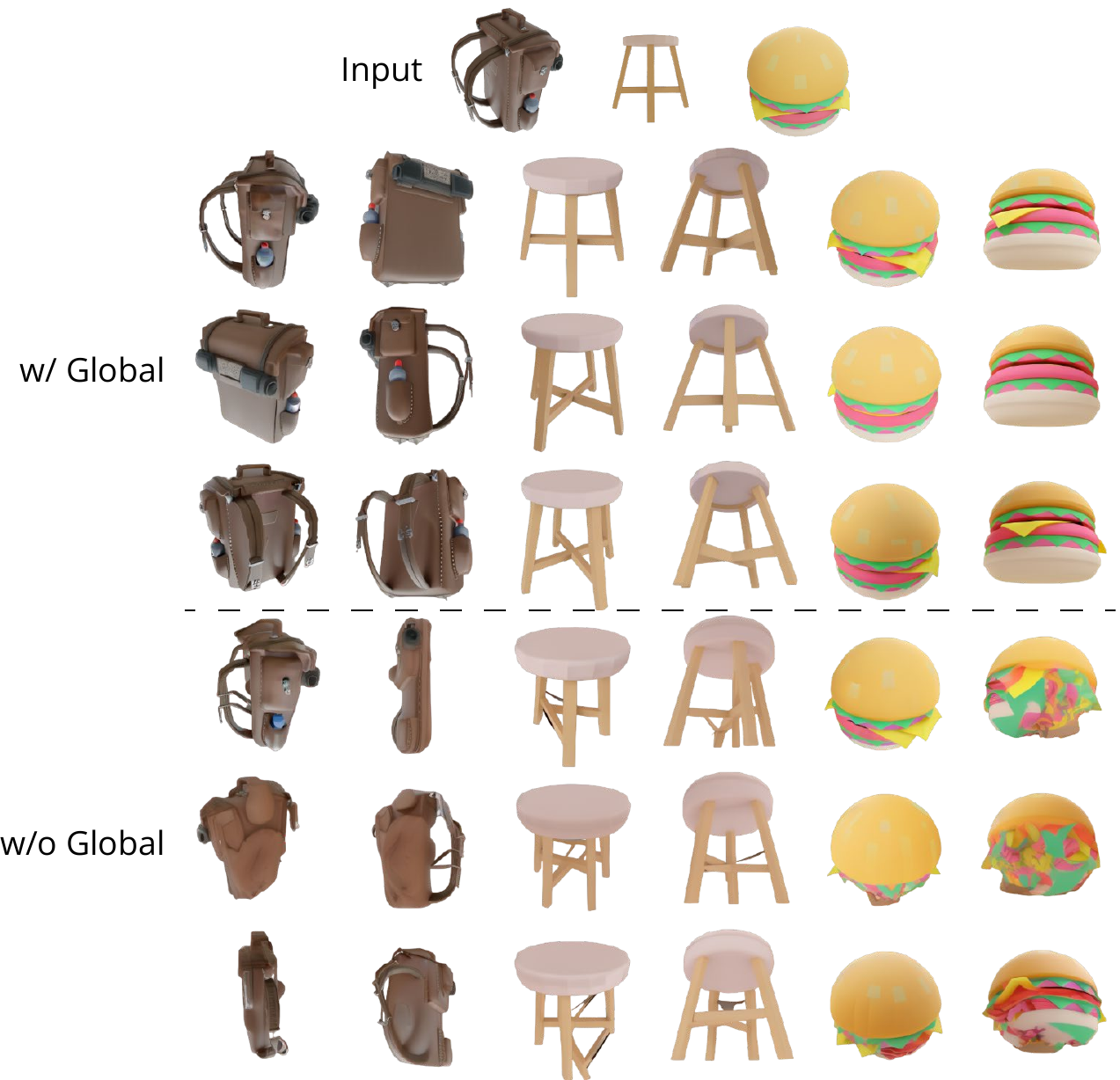}
    \caption{\textbf{Ablation on global conditioning.} In regions of the image that are not visible in the input, the results are significantly worse without global conditioning.}
    \label{fig:global}
\end{figure}

We present the results of trained Zero123++ models with and without global conditioning in Figure~\ref{fig:global}. In the absence of the proposed global conditioning, the quality of generated content remains satisfactory for visible regions corresponding to the input image. However, the generation quality significantly deteriorates for unseen regions, as the model lacks the ability to infer the global semantics of the object.

\subsection{Putting Everything Together}

Starting from the Stable Diffusion 2 $v$-model, we train our Zero123++ model using all the techniques mentioned above.
We train Zero123++ on Objaverse~\cite{deitke2023objaverse} data rendered with random HDRI environment lighting.

We adopt the phased training schedule from the Stable Diffusion Image Variations model~\cite{lambdaivd} to further reduce the extent of finetuning and preserve as much prior in Stable Diffusion as possible.
In the first phase, we only tune the self-attention layers and the KV matrices of cross-attention layers of Stable Diffusion. We use the AdamW\cite{kingma2014adam,loshchilov2017decoupled} optimizer with cosine annealing learning rate schedule peaking at $7 \times 10^{-5}$ and $1000$ warm-up steps.
In the second phase, we employ a very conservative constant learning rate of $5 \times 10^{-6}$ and $2000$ warm-up steps to tune the full UNet.
We employ the Min-SNR weighting strategy~\cite{hang2023efficient} to make the training process more efficient.

\section{Comparison to the State of the Art}

\subsection{Image to Multi-view}

\paragraph{Qualititative Comparison.} In Fig.~\ref{fig:i3d} we show generation results of 
% various methods including 
Zero-1-to-3 XL~\cite{liu2023zero,deitke2023objaversexl}, SyncDreamer~\cite{liu2023syncdreamer} and our Zero123++ on four input images, including one image from the Objaverse dataset with large uncertainty on the back side of the object (an electric toy cat), one real photo (extinguisher), one image generated by SDXL~\cite{podell2023sdxl} (a dog sitting on a rocket) and an anime illustration. We apply the elevation estimation method from One-2-3-45~\cite{liu2023one} for the required elevation estimation steps in Zero-1-to-3 XL and SyncDreamer. We use SAM~\cite{kirillov2023segment} for background removal. Zero123++ generates consistent and high-quality multi-view images, and can generalize to out-of-domain AI-generated and 2D illustration images.

\paragraph{Quantitative Comparison.} We evaluate the LPIPS score~\cite{zhang2018unreasonable} of different models on the validation split (subset of Objaverse) to quantitatively compare Zero-1-to-3~\cite{liu2023zero}, Zero-1-to-3 XL~\cite{liu2023zero,deitke2023objaversexl} and Zero123++. SyncDreamer~\cite{liu2023syncdreamer} is excluded because it does not support changing the elevation. To evaluate the multi-view generation results, we tile 6 generated images and the ground truth reference images (rendered from Objaverse) respectively, and compute the LPIPS score between the tiled images. Note that Zero-1-to-3 models may have seen our validation split during training, and the XL variant is trained on much more data than Zero123++. Nevertheless, Zero123++ achieves the best LPIPS score on the validation split. This shows the effectiveness of our designs in Zero123++. The results are shown in Tab.~\ref{tab:lpipsval}.

\begin{table}[tbh]
\centering
\caption{Quantitative results of models on our validation split.}
\label{tab:lpipsval}
\begin{tabular}{lc} 
\toprule
Model            & LPIPS $\downarrow$          \\ 
\midrule
Zero-1-to-3      & 0.210 $\pm$ 0.059          \\
Zero-1-to-3 XL & 0.188 $\pm$ 0.053          \\ 
\midrule
Zero123++ (Ours) & \textbf{0.177} $\pm$ 0.066  \\
\bottomrule
\end{tabular}
\end{table}

\begin{figure}
    \centering
    \includegraphics[width=0.95\linewidth]{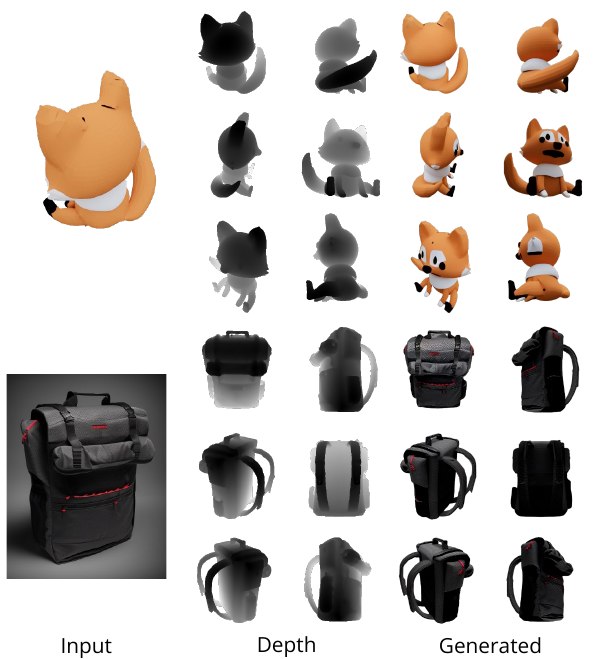}
    \caption{\textbf{Example outputs from depth-controlled Zero123++.} In the first example, we generate the face for the fox from its back; in the second example, we start with only the geometry and generate both the input image to Zero123++ and the multi-view images conditioned on the geometry.}
    \label{fig:depthc}
\end{figure}

\subsection{Text to Multi-view}

For text to multi-view, we first generate an image using SDXL with the text prompts, and then run Zero123++ upon the generated image.
In Fig.~\ref{fig:t3d}, we compare our results to MVDream~\cite{shi2023mvdream} and Zero-1-to-3 XL~\cite{liu2023zero,deitke2023objaversexl}.
We observe a texture style shift in MVDream to cartoonish and flat texture due to the bias in the Objaverse dataset, and the fact
that Zero-1-to-3 can not guarantee multi-view consistency,
while Zero123++ is able to generate realistic, consistent and highly detailed multi-view images using the text-to-image-to-multi-view pipeline.

\section{Depth ControlNet for Zero123++}

In addition to the base Zero123++ model, we also release a depth-controlled version of Zero123++ built with ControlNet~\cite{zhang2023adding}.
We render normalized linear depth images corresponding to the target RGB images and train a ControlNet to control Zero123++ on the geometry via depth.
The trained model is able to achieve a superior LPIPS of $0.086$ on our validation split.

Fig.~\ref{fig:depthc} shows two example generations from depth controlled Zero123++. We may use a single view as the input image to Zero123++ (the first example) or generate the input image from depth with vanilla depth-controlled Stable Diffusion as well to eliminate any need for input colors (the second example).

\begin{figure*}
    \centering
    \includegraphics[width=0.83\textwidth]{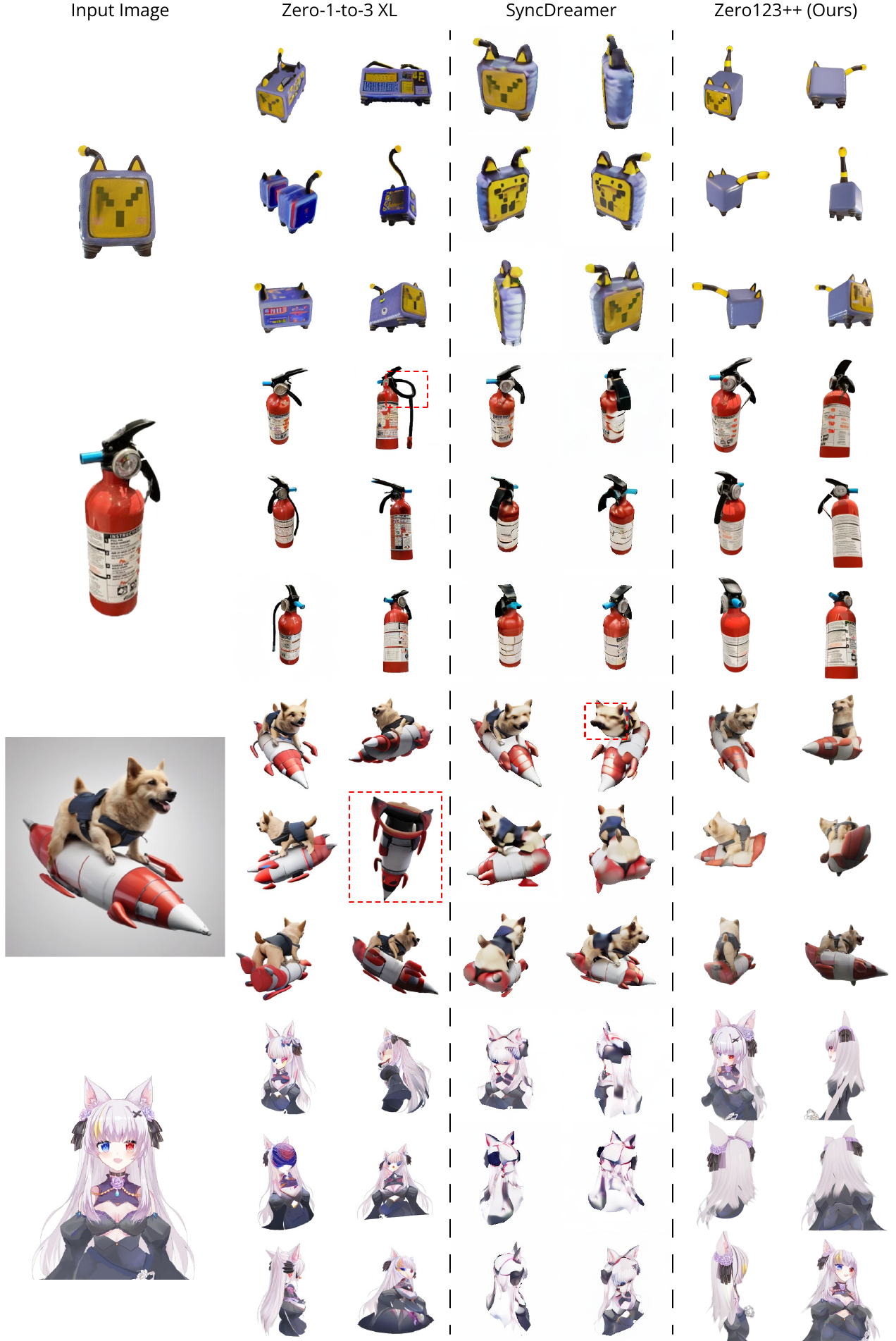}
    \caption{Qualitative comparison of Zero123++ against various methods on single image to multi-view task.}
    \label{fig:i3d}
\end{figure*}

\begin{figure*}[t]
    \centering
    \includegraphics[width=0.95\textwidth]{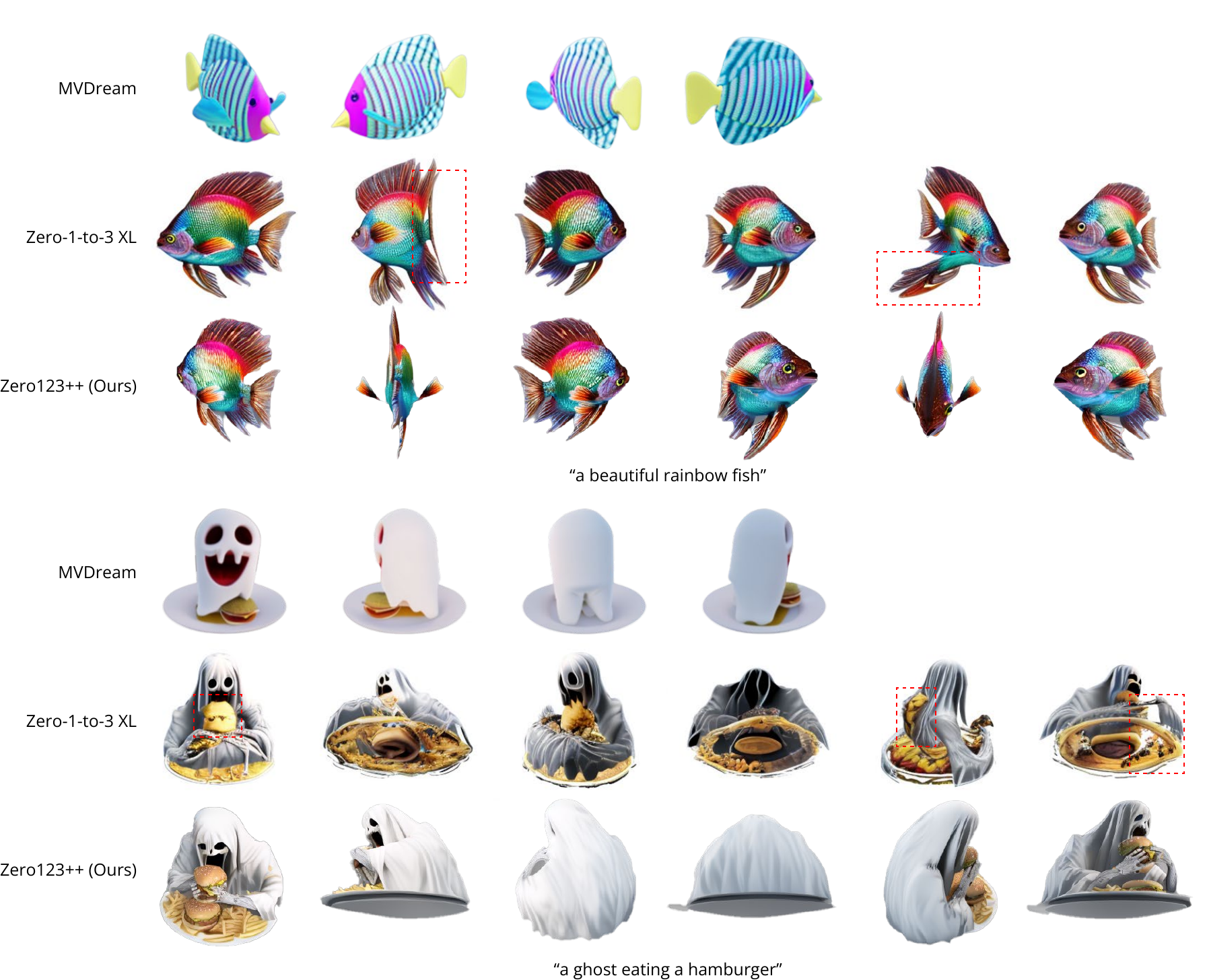}
    \caption{Qualitative comparison of Zero123++ against MVDream and Zero-1-to-3 XL on text to multi-view task.}
    \label{fig:t3d}
\end{figure*}

\begin{figure*}
    \centering
    \includegraphics[width=0.83\textwidth]{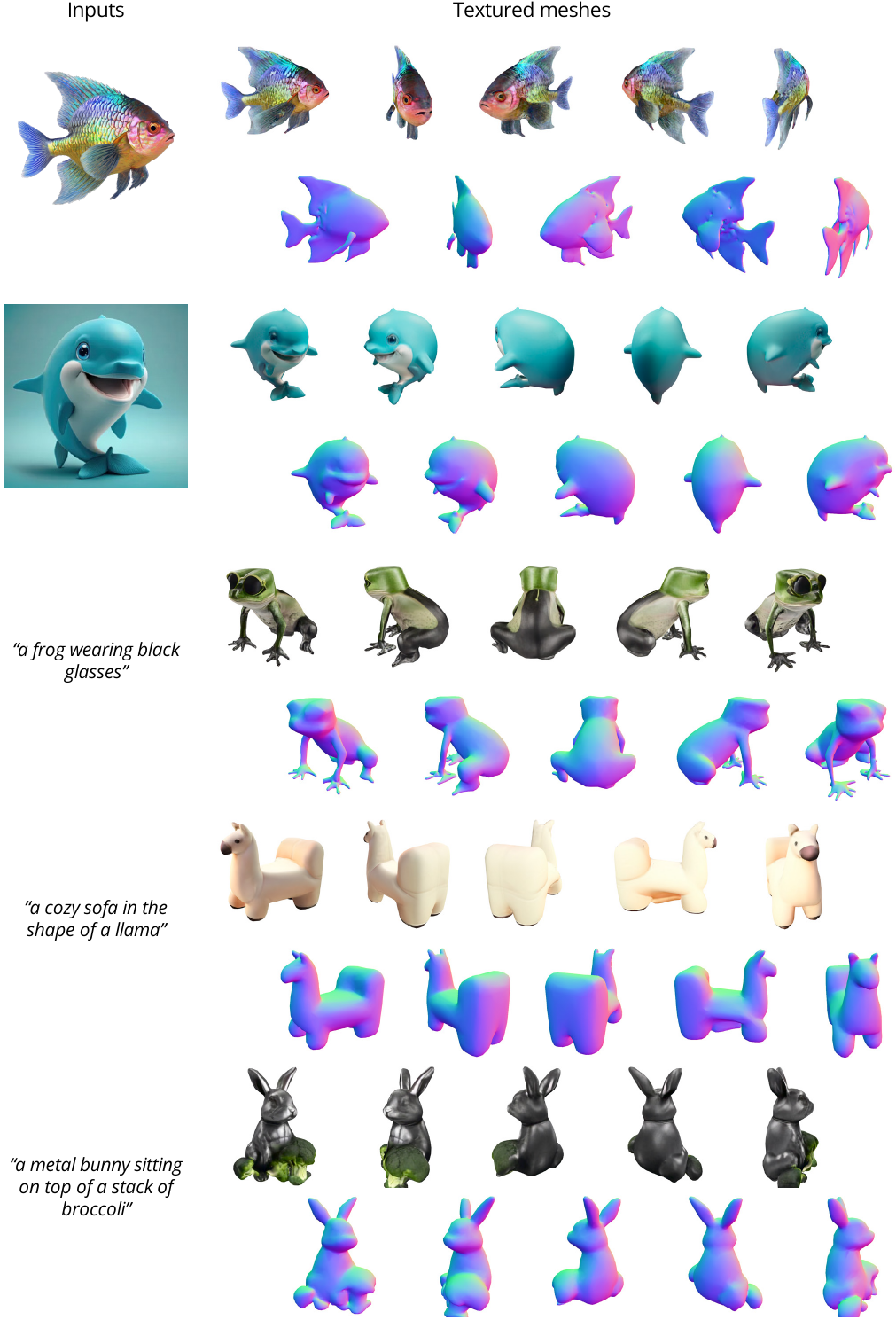}
    \caption{Preliminary mesh generation results with Zero123++.}
    \label{fig:mesh}
\end{figure*}

\section{Future Work}

This report presents an array of analyses and enhancements for our new image-to-multi-view base diffusion model, Zero123++. While our model has already achieved significantly improved quality, consistency, and generalization compared to previous models, we'd like to highlight three areas of potential future work:

\begin{itemize}
    \item Two-stage refiner model.
          Though $\epsilon$-parametrized models have trouble meeting the global requirements of consistency, they usually do better at generating local details.
          We may apply a two-stage generate-refine pipeline like SDXL~\cite{podell2023sdxl}, and employ the $\epsilon$-parametrized SDXL model as the base model for finetuning the refiner model, leveraging its stronger priors compared to the previous SD models.
    \item Further scaling-up.
          Currently Zero123++ is trained on the medium-scale Objaverse dataset, containing around 800k objects.
          To enhance our model's capabilities, we're looking into the possibility of scaling up our training to a larger dataset, such as Objaverse-XL~\cite{deitke2023objaversexl}.
    \item Utilizing Zero123++ for mesh reconstruction.
          There remains a gap between high quality multi-view images and high quality 3D meshes. We show some preliminary results utilizing Zero123++ for mesh generation in Fig.~\ref{fig:mesh}.
\end{itemize}

% \section*{Acknowledgements}

%%%%%%%%% REFERENCES
{\small
\bibliographystyle{ieee_fullname}
\bibliography{egbib}
}

\end{document}